\documentclass{article}

\usepackage{arxiv}

\usepackage[utf8]{inputenc} % allow utf-8 input
\usepackage[T1]{fontenc}    % use 8-bit T1 fonts
\usepackage{hyperref}   
\usepackage{tikz}
\usetikzlibrary{shapes.geometric, arrows}
\usepackage{url}            % simple URL typesetting
\usepackage{booktabs}       % professional-quality tables
\usepackage{amsfonts}       % blackboard math symbols
\usepackage{nicefrac}       % compact symbols for 1/2, etc.
\usepackage{microtype}      % microtypography
\usepackage{xcolor}         % colors
\usepackage{hyperref}
\usepackage{algorithm}
\usepackage{algpseudocode}
\usepackage{amsmath}
\usepackage{amssymb}
\usepackage{mathtools}
\usepackage{amsthm}% simple URL typesetting
\usepackage{graphicx}
\usepackage{caption}
\usepackage{subcaption}

\usepackage{comment}
\usepackage{wrapfig, blindtext}

\title{Permutation Decision Trees}

\author{%
   Harikrishnan N B\thanks{Adjunct Faculty, National Institute of Advanced Studies, IISc Campus, Karnataka, India, \\HNB webpage: https://sites.google.com/site/harikrishnannb8/home} \\
  Department of Computer Science \& Information Systems and APPCAIR\\
  BITS Pilani K K Birla Goa Campus, Goa, India\\
  \texttt{harikrishnannb@goa.bits-pilani.ac.in} \\
  \And
   Arham Jain \\
  Department of Computer Science \& Information Systems\\
  BITS Pilani K K Birla Goa Campus, Goa, India\\
  \texttt{f20211882@goa.bits-pilani.ac.in} \\
  \AND
Nithin Nagaraj \\
  Consciousness Studies Programme \\
  National Institute of Advanced Studies \\
  Indian Institute of Science Campus\\
  Bengaluru, Karnataka, India \\
  Email: \texttt{nithin@nias.res.in} \\
}

\begin{document}

\maketitle

\begin{abstract}

Decision Tree is a well understood Machine Learning model that is based on minimizing impurities in the internal nodes. The most common impurity measures are \emph{Shannon entropy} and \emph{Gini impurity}. These impurity measures are insensitive to the order of training data and hence the final tree obtained is invariant to any permutation of the data. This is a limitation in terms of modeling when there are temporal order dependencies between data instances. In this research, we propose the adoption of~\emph{Effort-To-Compress} (ETC) - a complexity measure, for the first time, as an alternative impurity measure.  Unlike Shannon entropy and Gini impurity, structural impurity based on ETC is able to capture order dependencies in the data, thus obtaining potentially different decision trees for different permutations of the same data instances, a concept we term as \emph{Permutation Decision Trees} (PDT). We then introduce the notion of {\it Permutation Bagging} achieved using permutation decision trees without the need for random feature selection and sub-sampling. We conduct a performance comparison between Permutation Decision Trees and classical decision trees across various real-world datasets, including \emph{Appendicitis}, \emph{Breast Cancer Wisconsin}, \emph{Diabetes Pima Indian}, \emph{Ionosphere}, \emph{Iris}, \emph{Sonar}, and \emph{Wine}. Our findings reveal that PDT demonstrates comparable performance to classical decision trees across most datasets. Remarkably, in certain instances, PDT even slightly surpasses the performance of classical decision trees. In comparing Permutation Bagging with Random Forest, we attain comparable performance to Random Forest models consisting of 50 to 1000 trees, using merely 21 trees. This highlights the efficiency and effectiveness of Permutation Bagging in achieving comparable performance outcomes with significantly fewer trees.
\end{abstract}

\section{Introduction}
The goal of Machine Learning (ML) is to develop algorithms that can learn from known data in order to predict outcomes for unseen future data. The success of ML models are typically measured in terms of their performance in accurately predicting the outcomes for the unseen data instances. Nevertheless, this evaluation remains incomplete without considering the interpretability of the model. Interpretability, or the ability to gather insights, plays a pivotal role in human comprehension and contributes significantly to establishing trustworthiness of the model. Some of the models, used in the category of interpretable ML are Decision Trees~\cite{quinlan1996learning}, Generalized Additive Models~\cite{hastie2017generalized}, and Decision Sets~\cite{lakkaraju2016interpretable}. 
The Decision Tree algorithm is widely recognized in the ML community. The interpretability of the tree is often assessed by its depth and the ability to trace back why a specific decision is favored over others. However, these advantages do not sufficiently explain why the Decision Tree algorithm, post-training, is interpretable for a domain expert. To enhance the interpretability of the model and align it with domain expertise, the insights provided by human domain experts should be incorporated into the final model before deploying it in actual practice.

\begin{wrapfigure}{H}{0.5\linewidth}
\vspace{-10pt}
\centering
\includegraphics[scale=0.55]{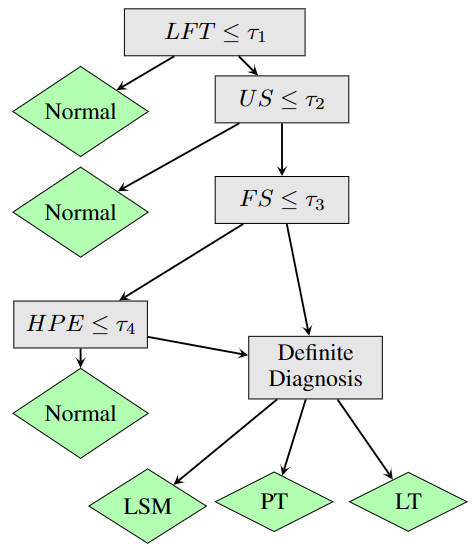}
%\rule{0.9\linewidth}{0.75\linewidth}
\caption{An interpretable decision tree designed to assess liver functionality and recommend appropriate remedies in the presence of detected abnormalities.}
\label{figure_dt_liver_health} 
 \vspace{-45pt}
  %\vspace{1pt}}
\end{wrapfigure}

In this research, the objective is to craft interpretable decision trees, employing the expertise of human domain experts in the final selection of the model(s). To underscore the significance of our work, we will employ practical examples, such as the examination of Liver cirrhosis, to elucidate the importance of our research.

\begin{enumerate}
    \item Liver Function Test (LFT)
    \item Ultrasound Scan (US) (if LFT is abnormal)
    \item Fibroscan (FS) (if US is abnormal, to grade the stiffness)
    \item Histopathological examination (HPE) (to check for localized lesion)
    \item Definite Diagnosis: (a) Life Style Modification (LSM), (b) Pharmacological Treatment (PT), and (c) Liver Transplantation (LT)
\end{enumerate}

Liver cirrhosis stands as a significant contributor to morbidity and mortality among individuals with chronic liver disease globally. In $2019$, cirrhosis accounted for $2.4\%$ of global deaths~\cite{huang2023global}. Various tests are available to ascertain the health of the liver and determine whether it falls within the healthy range. According to medical domain experts, the sequential series of tests to evaluate liver health are as follows:

The aforementioned sequence represents the commonly recommended order of tests by medical experts for evaluating the condition of liver health. The corresponding decision tree diagram illustrating these tests is presented in Figure~\ref{figure_dt_liver_health}.

An individual taking these test will be classified to the following categories:
\begin{enumerate}
    \item Normal
    \item  Life Style Modification 
    \item Pharmacological Treatment 
    \item Liver Transplantation 
    
\end{enumerate}
The decision tree provided in Figure~\ref{figure_dt_liver_health} is the sequence of steps followed by a medical expert. 
This problem of evaluating the liver health condition can be mapped to Machine Learning paradigm. The problem can be seen as a four class classification problem with the classes as follows: $Normal$, $LSM$, $PT$ and $LT$. 
In the machine learning paradigm, our training dataset comprises $m$ instances, each characterized by $n$ features, namely outcome results/values of $LFT$, $US$, $FS$, and $HPE$. As previously outlined, the problem can be framed as a four class classification challenge. The task involves categorizing a new patient based on their medical history into one of these four classes. The use of the aforementioned decision tree ensures interpretable results, as its design is informed by domain expertise from human experts. However, within the machine learning paradigm, there is no assurance that constructing a decision tree will inherently adhere to the specific test order presented in Figure~\ref{figure_dt_liver_health}.This is because Decision Tree uses either Shannon Entropy or Gini impurity as an impurity measure so as to maximize the information gain. The goal of Decision Trees in the ML context is to maximize the information gain. Maximizing information gain does not guarantee a decision tree that is interpretable for a domain expert. In the given example, suppose the Decision Tree suggests starting with a test like Fibroscan followed by Histopathological examination, deviating from the recommended LFT initiation. This could lead to impractical outcomes, such as recommending costly tests that may not be universally accessible in local hospitals, thus hindering patient-friendliness. So there arise a need to have human in the loop for final model selection. The proposed solution in this research is the implementation of ``Permutation Decision Trees" (PDT). PDT generates multiple decision trees based on various permutations of data instances, offering a diverse set of models. In contrast, Decision Trees using Shannon Entropy or Gini impurity typically yield a single decision tree. The set of distinct Decision Trees provided by PDT facilitates human domain experts in selecting a suitable model or multiple models for deployment, enhancing interpretability and practicality in real-world applications. 

\section{Related Works}
In our research, we introduce the novel use of `Effort-To-Compress' (ETC) -- a compression-complexity measure -- as an {\emph{impurity function} for Decision Trees, marking the first instance of its application in Machine Learning. ETC effectively measures the effort required for lossless compression of an object through a predetermined lossless compression algorithm~\cite{nagaraj2017three}. ETC was initially introduced in~\cite{nagaraj2013new} as a measure of complexity for timeseries analysis, aiming to overcome the severe limitations of entropy-based complexity measures. It is worth noting that the concept of complexity lacks a singular, universally accepted definition. In~\cite{nagaraj2017three}, complexity was explored from different perspectives, including the effort-to-describe (Shannon entropy, Lempel-Ziv complexity), effort-to-compress (ETC complexity), and degree-of-order (Subsymmetry). The same paper highlighted the superior performance of ETC in distinguishing between periodic and chaotic timeseries. Moreover, ETC has played a pivotal role in the development of an interventional causality testing method known as Compression-Complexity-Causality (CCC)~\cite{kathpalia2019data}. CCC and allied approaches based on compression-complexity  have been rigorously tested in several practical applications for causal discovery and inference~\cite{pranay2021causal, ramanan2022detection, kathpalia2022compression, nb2022causality}. ETC has demonstrated robust, reliable and superior performance over infotheoretic approaches when applied to short and noisy time series data (including stochastic and/or chaotic ones), leading to its utilization in diverse fields such as investigating cardiovascular dynamics~\cite{balasubramanian2017vagus}, conducting cognitive research~\cite{kimiskidis2015transcranial}, and analysis of muscial compositions~\cite{nandekar2021causal}.
\section{Proposed Method}
In this section, we establish the concept of structural impurity and subsequently present an illustrative example to aid in comprehending the functionality of ETC.

\emph{Definition:} Structural impurity for a sequence $S = s_1, \ldots, s_n$, where $s_i \in \{ 1,2\ldots, K\}$, and $K \in \mathbf{Z^{+}}$ is the  extent of \emph{irregularity} in the sequence $S$.

The above definition is purposefully generic in its use of the term \emph{irregularity} and one could use several specific measures of \emph{irregularity}. We suggest a specific measure, namely `Effort-To-Compress' or ETC for measuring the extent of irregularity in the input sequence $S$ and thus inturn serving as a measure of structual impurity.ETC algorithm is defined in the Appendix section~\ref{sec_ETC_algorithm}. We will now illustrate how ETC serves as a measure of structural impurity. The formal definition of ETC is the effort required for the lossless compression of an object using a predefined lossless compression algorithm.  The specific algorithm employed to compute ETC is known as Non-sequential Recursive Pair Substitution (NSRPS). NSRPS was initially proposed by Ebeling~\cite{ebeling1980grammars} in 1980 and has since undergone improvements~\cite{jimenez2002entropy}, ultimately proving to be an optimal choice~\cite{benedetto2006non}. Notably, NSRPS has been extensively utilized to estimate the entropy of written English~\cite{grassberger2002data}. The algorithm is briefly discussed below:
Let's consider the sequence $S = 11122$ to demonstrate the iterative steps of the algorithm. In each iteration, we identify the pair of symbols with the highest frequency (algorithm~\ref{alg_pair_wise_max}-for finding the most frequently occurring pairs in the given sequence; algorithm~\ref{alg_first_occuring_max_pair}-from the most frequently occurring pairs, choose the pair that comes first in the given sequence; this pair is replaced by a new symbol) and replace all non-overlapping instances of that pair with a new symbol (algorithm~\ref{alg_ETC_algorithm} - replace the most frequently occurring pair by a new symbol and eventually compute the ETC value). In the case of sequence $S$, the pair with the maximum occurrence is $11$. We substitute all occurrences of $11$ with a new symbol, let's say $3$, resulting in the transformed sequence $3122$. We continue applying the algorithm iteratively. The sequence $3122$ is further modified to become $422$, where the pair $31$ is replaced by $4$. Then, the sequence $422$ is transformed into $52$ by replacing $42$ with $5$. Finally, the sequence $52$ is substituted with $6$. At this point, the algorithm terminates as the stopping criterion is achieved when the sequence becomes homogeneous. ETC, as defined in~\cite{nagaraj2013new}, is equal to the count of the number of iterations needed for the NSRPS algorithm to attain a homogeneous sequence when applied iteratively on the input sequence (as described above -algorithm~\ref{alg_ETC_algorithm}).
\subsection{ETC as an impurity measure}

We consider the structural impurity (computed using ETC) of the following binary sequences and compare it with Shannon entropy and Gini impurity measures.
\begin{table*}[h!]
\centering
\caption{Comparison of Structural impurity computed using ETC with Shannon entropy, and Gini impurity for various binary sequences.}
\scalebox{0.8}{
\begin{tabular}{|c|c|c|c|c|}
\hline
\textbf{Sequence ID} &\textbf{Sequence} & \textbf{Structural Impurity (ETC)} & \textbf{Entropy} & \textbf{Gini Impurity} \\ \hline
A & 111111 & 0                    & 0                   & 0          \\ \hline
B & 121212   & 1 & 1 & 0.5 \\ \hline
C & 222111 & 5 & 1 & 0.5 \\ \hline
D & 122112 & 4 & 1 & 0.5 \\ \hline
E & 211122 & 5 & 1 & 0.5 \\ \hline

\end{tabular}
}
\label{table_etc_example}
\end{table*}

Referring to Table~\ref{table_etc_example}, we observe that for sequence A, the ETC, Shannon Entropy, and Gini impurity all have a value of zero. This outcome arises from the fact that the sequence is homogeneous, devoid of any impurity. However, for sequences B, C, D, and E, the Shannon entropy and Gini impurity remain constant, whereas ETC varies based on the particular {\it structural characteristics} of each sequence. Thus, ETC is able to capture the {\it structural impurity} in sequences, better than both Shannon entropy and Gini impurity measures. 
\subsection{Relationship between Average ETC and Shannon Entropy}
In this section, we study the relationship between ETC and Shannon Entropy. We consider all permutations of binary array of length ($n=20$) with number of zeros ($k$) in the array varying from $0$ (implying all entries are one) to $n$. We compute the mean Normalized ETC values (Normalized $ETC_{mean}$) and mean Shannon Entropy for all the permutations of binary string corresponding to $k$ varying from $0$ to $n$. (Note: Shannon entropy of a fixed $k$ value is invariant to permutation of the binary array). The Normalized $ETC(x)$ of a binary string $x$ of length $n$ is defined as $\frac{ETC(x)}{n-1}$. For the following setting: $n=20$, $k=0$ to $n$ with a step size of $1$, we obtain a positively correlated relationship between Normalized $ETC_{mean}$ and Shannon Entropy. Figure~\ref{figure_H_Normalized_ETC_mean_vs_num_zeros} depicts this relationship. The symmetric behavior found in average ETC can be explained as follows. For an $n$-length binary string with $k$ zeros, the total permutations possible are $\binom{n}{k}$. Similarly, if the number of zeros is $n-k$, then the total permutations possible are $\binom{n}{n-k}$. But we know that $\binom{n}{n-k} = \binom{n}{k}$, implying that the total number and structure of binary strings are the same (the difference is only in the alphabets). Since ETC looks at the structure and not the numeric representation, the average ETC will be the same for $k$ zeros and $n-k$ zeros leading to a symmetry in the relationship between (Normalized $ETC_{mean}$) and $k$ as depicted in Figure~\ref{figure_H_Normalized_ETC_mean_vs_num_zeros}. Interestingly, it has been shown that expected \emph{Kolmogrov complexity} equals Shannon entropy~\cite{grunwald2004shannon}. In future work, we aim to explore potential connections between ETC and Kolmogorov complexity

\begin{figure}[!ht]
% \centerline{\includegraphics[width=8cm,height=5.0cm]{method1.png}}
%\centerline{ \includegraphics[width=0.5\textwidth]{H_Normalized_ETC_mean_vs_num_zeros.png}}
\centerline{ 
\includegraphics[scale=0.21]{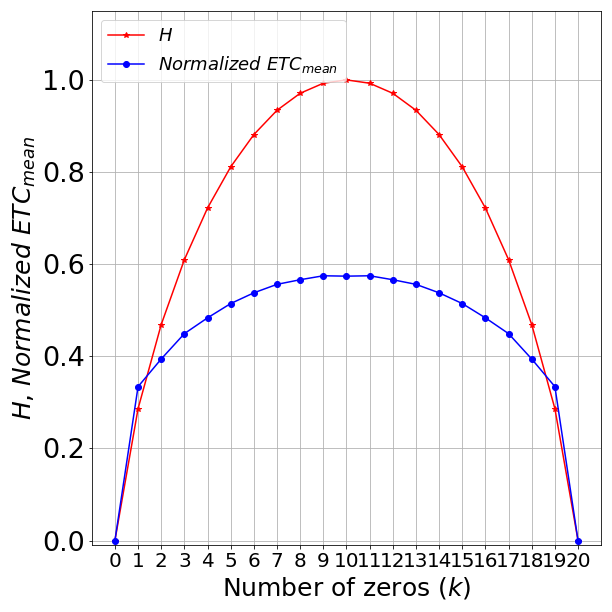}
\includegraphics[scale=0.4]{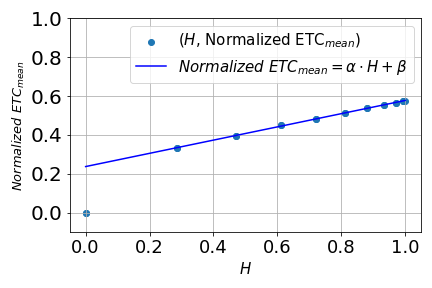}
}
\caption{(a) Relationship between Normalized ETC mean and Shannon Entropy with respect to permutations of the binary string (of length $n=20$) with number of zeros $k$ varying from $0$ to $20$. (b) Relationship between Normalized ETC mean with respect to Shannon Entropy for permutations of the binary string (of length $20$) with number of zeros $k$ varying from $1$ to $19$. In the plot, we excluded the case where the binary string is completely a zero entry or a ones entry array. }
\label{figure_H_Normalized_ETC_mean_vs_num_zeros}
\end{figure}

To further understand the relationship between Normalized $ETC_{mean}$ and Shannon Entropy, we performed a linear regression between Normalized $ETC_{mean}$ (as the $y-axis$) and Shannon Entropy (as the x-axis) for $k$ varying from $1$ to $n-1$. We exclude the case where the binary string is completely a zero entry or a ones entry array. For the given setting of $n$ and $k$, we found a positively correlated linear relationship between Normalized $ETC_{mean}$ and Shannon Entropy ($H$). Figure~\ref{figure_H_Normalized_ETC_mean_vs_num_zeros} (a) depicts the aforementioned relationship. We found that for $n=20$, Normlaized $ETC_{mean} = \alpha*H +\beta$, where $\alpha = 0.340$\footnote{0.34007669} and $\beta = 0.237$\footnote{0.23756388}.

\begin{comment}
  \begin{figure}[!ht]
% \centerline{\includegraphics[width=8cm,height=5.0cm]{method1.png}}
\centerline{ \includegraphics[width=0.5\textwidth]{Normalized_ETC_mean_vs_H.png}}
\caption{Relationship between Normalized ETC mean with respect to Shannon Entropy for permutations of the binary string (of length $20$) with number of zeros $k$ varying from $1$ to $19$. In the plot, we excluded the case where the binary string is completely a zero entry or a ones entry array.}
\label{figure_Normalized_ETC_mean_vs_H}
\end{figure}
\end{comment}

This indicates that Normalized $ETC_{mean}$/ $ETC_{mean}$/ ETC very well can be used as an impurity measure leading to a decision tree consistent with traditional decision tree. However, ETC also provides flexibility in allowing certain permutations and developing a decision tree which is variant from traditional decision tree. 
\subsection{ETC Gain}
Having shown that the ETC captures the structural impurity of a sequence, we now define \emph{ETC Gain}. ETC gain is the reduction in ETC caused by partitioning the data instances according to a particular attribute of the dataset. Consider the decision tree structure provided in Figure~\ref{figure_etc_gain}. 

% \begin{figure}[!ht]
%     \centering
% \begin{center}
% \tikzset{
%   block/.style = {rectangle, draw, fill=gray!20,
%                   text width=#1, minimum height=7mm, align=flush center},
%  block/.default = 5 em,
% decision/.style = {diamond, aspect=1.5, draw, fill=green!30,
%                    minimum width=5em,align=center}
%         }\tikzstyle{arrow} = [thick,->,>=stealth]
%     \begin{tikzpicture}[
% node distance = 5mm and 5mm,
%                         ]
% \node[block=7em]       (A) {Parent};
% \node[block,below=of A, xshift=-1cm,yshift=-0.8cm] (B) {Left Child};
% \node[block,right=of B,xshift=0.5cm,yshift=-1.15cm] (C) {Right Child};

% % \node[decision, below=of D, xshift=2cm,yshift=-2.7cm] (H) {YES};
% % \node[decision, left=of H] (I) {NO};
% \draw [arrow] (A) -- (B);
% \draw [arrow] (A) -- (C);

%     \end{tikzpicture}
%    \caption{Decision Tree structure with a parent node and two child node (Left Child and Right Child).}
%     \label{figure_etc_gain}   
% \end{center}

% \end{figure}

\begin{figure}[!ht]

\centering
\includegraphics[scale=0.55]{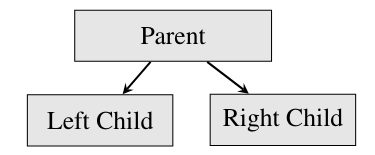}
%\rule{0.9\linewidth}{0.75\linewidth}
\caption{Decision Tree structure with a parent node and two child node (Left Child and Right Child).}
    \label{figure_etc_gain}    

  %\vspace{1pt}}
\end{figure}

The ETC Gain between the chosen parent attribute ($P$) of the tree and the class labels ($S$) is defined as follows:
\begin{equation}
    ETC~Gain(S, P) = ETC(S)-\displaystyle\sum\limits_{V\in Values(P)}\frac{|S_{V}|}{|S|}\cdot ETC(S_V),
    \label{equation_etc_gain}
\end{equation}

%%%%%%%%%%%%%%%
where $V$ in equation~\ref{equation_etc_gain} is the possible values of the chosen parent attribute $P$, $S_V \subset S$, and $S_V$ consists of labels corresponding to the chosen parent attribute $P$ taking the value $V$, $|S|$ represents the cardinality of the set $S$. The formula for ETC Gain, as given in equation~\ref{equation_etc_gain}, bears resemblance to information gain. The key distinction lies in the use of ETC as a measure of structural impurity instead of Shannon entropy in the calculation.

We use ETC as an impurity measure in the \emph{Permutation Decision Tree} (PDT) algorithm.
%%%%%%%%%%%%%%%
\subsection{Toy Example to illustrate the working of PDT }
To showcase the effectiveness of the proposed structural impurity measure (computed using ETC) in capturing the underlying structural dependencies within the data and subsequently generating distinct decision trees for different permutations of input data, we utilize the following illustrative toy example.
%%%%%%%%%

\begin{table}[h!]
\centering
\caption{Toy example dataset to demonstrate permuted decision trees generated with a novel structural impurity measure (computed using ETC). The toy example considered is a binary classification problem (class-1, class-2) with two features $f_0$, $f_1$ and $14$ data instance. The rows represents different training data instances.}
\scalebox{0.8}{
\begin{tabular}{|c|c|c|c|}
\hline
\textbf{Serial No.} & \textbf{$f_0$} & \textbf{$f_1$} & \textbf{label} \\ \hline
1 & 1                     & 1                     & 2              \\ \hline
2 & 1                     & 2                     & 2              \\ \hline
3 & 1                     & 3                     & 2              \\ \hline
4 & 2                     & 1                     & 2              \\ \hline
5 & 2                     & 2                     & 2              \\ \hline
6 & 2                     & 3                     & 2              \\ \hline
7 & 4                     & 1                     & 2              \\ \hline
8 & 4                     & 2                     & 2              \\ \hline
9 & 4                     & 3                     & 1              \\ \hline
10 & 4                     & 4                     & 1              \\ \hline
11 & 5                     & 1                     & 1              \\ \hline
12 & 5                     & 2                     & 1              \\ \hline
13 & 5                     & 3                     & 1              \\ \hline
14 & 5                     & 4                    & 1              \\ \hline
\end{tabular}
}
\label{table_toy_example}
\end{table}

The visual representation of the toy example provided in Table~\ref{table_toy_example} is represented in Figure~\ref{fig_toy_example}.
\begin{figure}[!h]
%\centerline{ \includegraphics[width=0.5\textwidth]{toy_example.png}}
\centerline{ 
\includegraphics[scale=0.3]{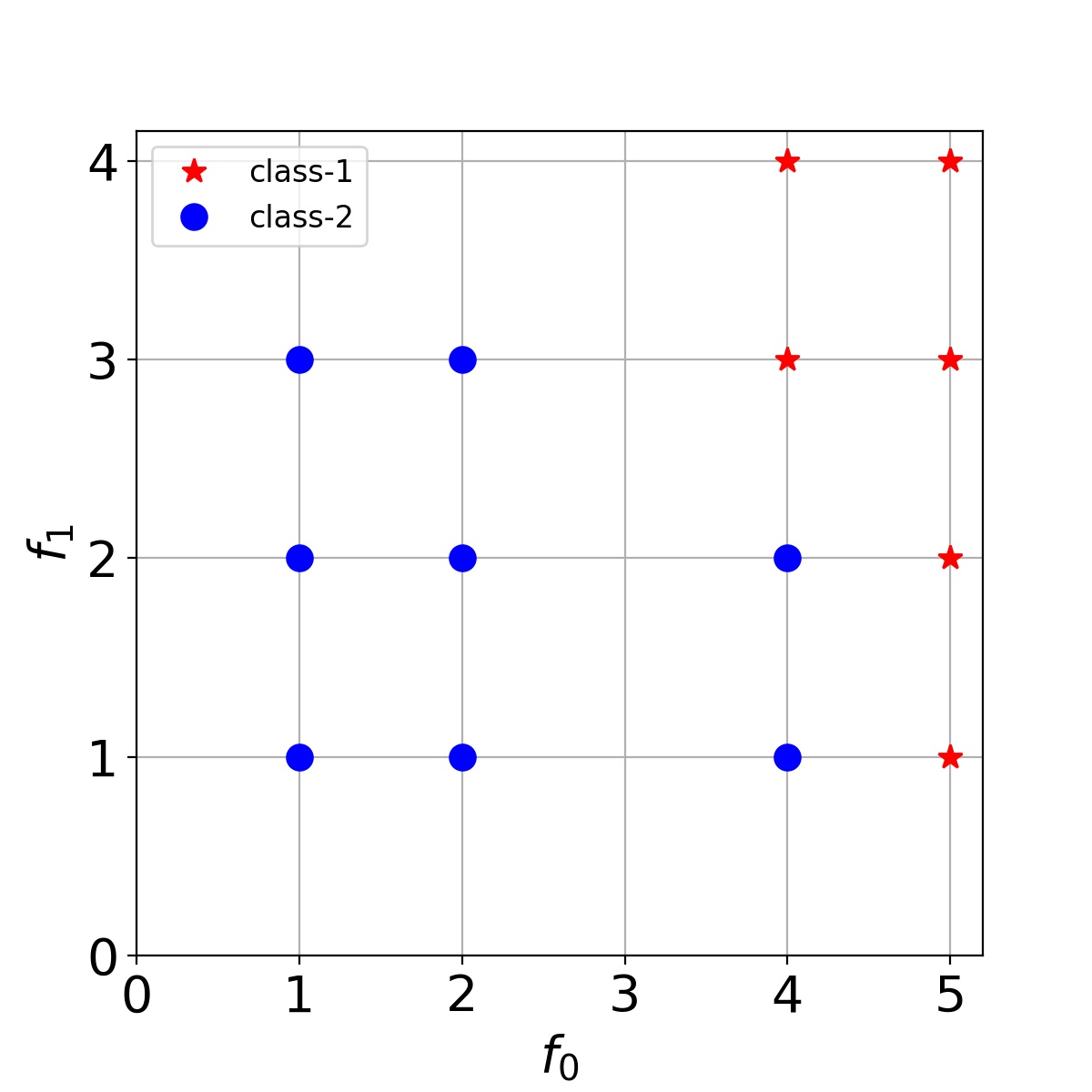}
\includegraphics[scale=0.5]{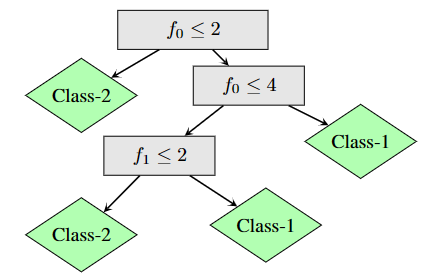}
}
\caption{(a) Left: A visual representation of the toy example provided in Table~\ref{table_toy_example}. (b) Right: Decision tree using the proposed structural impurity (computed using ETC) for Permutation A.}
\label{fig_toy_example}
\end{figure}
%%%%%%%%%
%
We consider the following permutations of the dataset. For each of the below permutations, we obtain distinct decision trees using the proposed structural impurity measure (computed using ETC).

\begin{itemize}
    \item Permutation A: 1, 2, 3, 4, 5, 6, 7, 8, 9, 10, 11, 12, 13, 14. Figure~\ref{fig_toy_example}(b) represents the corresponding decision tree.

%%%%%%%%%%%%%%%
\begin{comment}
\begin{figure}[!ht]
    \centering
\begin{center}
\tikzset{
  block/.style = {rectangle, draw, fill=gray!20,
                  text width=#1, minimum height=7mm, align=flush center},
 block/.default = 5 em,
decision/.style = {diamond, aspect=1.5, draw, fill=green!30,
                   minimum width=5em,align=center}
        }\tikzstyle{arrow} = [thick,->,>=stealth]
    \begin{tikzpicture}[
node distance = 5mm and 5mm,
                        ]
\node[block=7em]       (A) {$f_0 \leq 2$};
\node[decision,below=of A, xshift=-2cm, yshift=-0.5cm] (B) {Class-2};
\node[block,right=of B, yshift=-1cm] (C) {$f_0\leq 4$};
\node[block, below=of C, xshift=-0.6cm, yshift=-1.9cm] (D) {$f_1 \leq 2$};
\node[decision,right=of D, xshift=2cm, yshift=-2cm] (E) {Class-1};
\node[decision, below=of D, xshift=-2cm, yshift=-3cm] (F) {Class-2};
\node[decision, right=of F, xshift=0.3cm, yshift=-3.5cm] (G) {Class-1};
% \node[decision, below=of D, xshift=2cm,yshift=-2.7cm] (H) {YES};
% \node[decision, left=of H] (I) {NO};
\draw [arrow] (A) -- (B);
\draw [arrow] (A) -- (C);
\draw [arrow] (C) -- (D);
\draw [arrow] (C) -- (E);
\draw [arrow] (D) -- (F);
\draw [arrow] (D) -- (G);

    \end{tikzpicture}
   \caption{Decision tree using the proposed structural impurity (computed using ETC) for Permutation A.}
    \label{figure_dt_sequence_A}   
\end{center}
\end{figure}
\end{comment}

%%%%%%%%%%%%%%%
    \item  Permutation B: 14, 3, 10, 12, 2, 4, 5, 11, 9, 8, 7, 1, 6, 13. Figure~\ref{figure_dt_sequenceBC}(a) represents the corresponding decision tree.

%%%%%%%%%%%%%%%
\begin{comment}
\begin{figure}[!ht]
    \centering
\begin{center}
\tikzset{
  block/.style = {rectangle, draw, fill=gray!20,
                  text width=#1, minimum height=7mm, align=flush center},
 block/.default = 5 em,
decision/.style = {diamond, aspect=1.5, draw, fill=green!30,
                   minimum width=5em,align=center}
        }\tikzstyle{arrow} = [thick,->,>=stealth]
    \begin{tikzpicture}[
node distance = 5mm and 5mm,
                        ]
\node[block=7em]       (A) {$f_1 \leq 2$};
\node[block,below=of A, xshift=-2cm, yshift=-0.5cm] (B) {$f_0 \leq 4$};
\node[decision,below=of B, xshift=-4cm,yshift=-1.4cm] (C) {Class-2};
\node[decision, right=of C, xshift=-2cm, yshift=-2cm] (D) {Class-1};
\node[block,right=of B, xshift=1.2cm, yshift=-0.9cm] (E) {$f_0\leq 2$};
\node[decision,below=of E, xshift=1cm, yshift=-1.5cm] (F) {Class-2};
\node[decision, right=of F, xshift=4cm, yshift=-1.9cm] (G) {Class-1};

% \node[decision, below=of D, xshift=2cm,yshift=-2.7cm] (H) {YES};
% \node[decision, left=of H] (I) {NO};
\draw [arrow] (A) -- (B);
\draw [arrow] (A) -- (E);
\draw [arrow] (B) -- (C);
\draw [arrow] (B) -- (D);
\draw [arrow] (E) -- (F);
\draw [arrow] (E) -- (G);

    \end{tikzpicture}
   \caption{Decision tree using the proposed structural impurity (computed using ETC) for Permutation B.}
    \label{figure_dt_sequence_B}   
\end{center}

\end{figure}
\end{comment}

%%%%%%%%%%%%%%%

    \item Permutation C: 13, 11, 8, 12, 7, 6, 4, 14, 10, 5, 2, 3, 1, 9. Figure~\ref{figure_dt_sequenceBC}(b) represents the corresponding decision tree.

%%%%%%%%%%%%%%%

\begin{figure}[!h]
\centerline{ \includegraphics[scale=0.6]{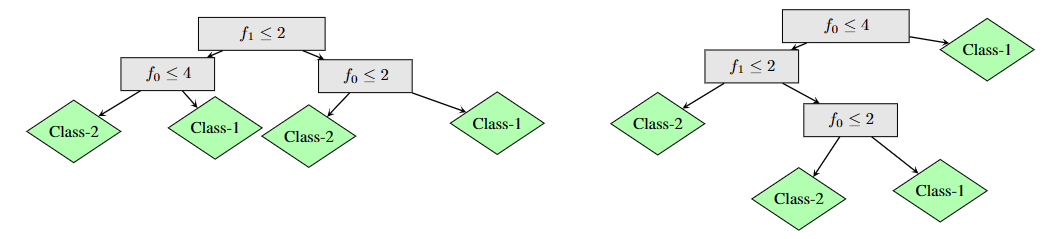}}
\caption{Decision tree using the proposed structural impurity (computed using ETC) for (a) Left: Permutation B, (b) Permutation C.}
\label{figure_dt_sequenceBC} 
\end{figure}

\begin{comment}
\begin{figure}[!ht]
    \centering
\begin{center}
\tikzset{
  block/.style = {rectangle, draw, fill=gray!20,
                  text width=#1, minimum height=7mm, align=flush center},
 block/.default = 5 em,
decision/.style = {diamond, aspect=1.5, draw, fill=green!30,
                   minimum width=5em,align=center}
        }\tikzstyle{arrow} = [thick,->,>=stealth]
    \begin{tikzpicture}[
node distance = 5mm and 5mm,
                        ]
\node[block=7em]       (A) {$f_0 \leq 4$};
\node[block,below=of A, xshift=-2cm, yshift=-0.5cm] (B) {$f_1 \leq 2$};
\node[decision,below=of B, xshift=-4cm,yshift=-1.4cm] (C) {Class-2};
\node[decision, right=of B,xshift=2cm, yshift=-0.5cm] (D) {Class-1};
\node[block,right=of C,xshift=-0.9cm, yshift=-2cm] (E) {$f_0 \leq 2$};
\node[decision, below=of E, xshift=-1cm, yshift=-3cm] (F) {Class-2};
\node[decision, right=of F,xshift=1cm, yshift=-3.5cm] (G) {Class-1};
% \node[decision, below=of D, xshift=2cm,yshift=-2.7cm] (H) {YES};
% \node[decision, left=of H] (I) {NO};
\draw [arrow] (A) -- (B);
\draw [arrow] (A) -- (D);
\draw [arrow] (B) -- (C);
\draw [arrow] (B) -- (E);
\draw [arrow] (E) -- (F);
\draw [arrow] (E) -- (G);

    \end{tikzpicture}
   \caption{Decision tree using the proposed structural impurity (computed using ETC) for Permutation C.}
    \label{figure_dt_sequence_C}   
\end{center}

\end{figure}
\end{comment}
%%%%%%%%%%%%%%%
    
    \item Permutation D: 3, 2, 13, 10, 11, 1, 4, 7, 6, 9, 8, 14, 5, 12. Figure~\ref{figure_dt_sequenceDE}(a) represents the corresponding decision tree.

%%%%%%%%%%%%%%%
\begin{comment}
\begin{figure}[!ht]
    \centering
\begin{center}
\tikzset{
  block/.style = {rectangle, draw, fill=gray!20,
                  text width=#1, minimum height=7mm, align=flush center},
 block/.default = 5 em,
decision/.style = {diamond, aspect=1.5, draw, fill=green!30,
                   minimum width=5em,align=center}
        }\tikzstyle{arrow} = [thick,->,>=stealth]
    \begin{tikzpicture}[
node distance = 5mm and 5mm,
                        ]
\node[block=7em]       (A) {$f_0 \leq 4$};
\node[block,below=of A, xshift=-2cm, yshift=-0.5cm] (B) {$f_0 \leq 2$};
\node[decision,below=of B, xshift=-4cm,yshift=-1.4cm] (C) {Class-2};
\node[decision, right=of B,xshift=2cm, yshift=-0.5cm] (D) {Class-1};
\node[block,right=of C,xshift=-0.9cm, yshift=-2cm] (E) {$f_1 \leq 2$};
\node[decision, below=of E, xshift=-1cm, yshift=-3cm] (F) {Class-2};
\node[decision, right=of F,xshift=1cm, yshift=-3.5cm] (G) {Class-1};
% \node[decision, below=of D, xshift=2cm,yshift=-2.7cm] (H) {YES};
% \node[decision, left=of H] (I) {NO};
\draw [arrow] (A) -- (B);
\draw [arrow] (A) -- (D);
\draw [arrow] (B) -- (C);
\draw [arrow] (B) -- (E);
\draw [arrow] (E) -- (F);
\draw [arrow] (E) -- (G);

    \end{tikzpicture}
   \caption{Decision tree using the proposed structural impurity (computed using ETC) for Permutation D.}
    \label{figure_dt_sequence_D}   
\end{center}

\end{figure}
\end{comment}
%%%%%%%%%%%%%%%
    
    \item Permutation E: 10, 12, 1, 2, 13, 14, 8, 11, 4, 7, 9, 6, 5, 3. Figure~\ref{figure_dt_sequenceDE}(b) represents the corresponding decision tree.

    %%%%%%%%%%%%%%%
\begin{comment}
       
\begin{figure}[!ht]
    \centering
\begin{center}
\tikzset{
  block/.style = {rectangle, draw, fill=gray!20,
                  text width=#1, minimum height=7mm, align=flush center},
 block/.default = 5 em,
decision/.style = {diamond, aspect=1.5, draw, fill=green!30,
                   minimum width=5em,align=center}
        }\tikzstyle{arrow} = [thick,->,>=stealth]
    \begin{tikzpicture}[
node distance = 5mm and 5mm,
                        ]
\node[block=7em]       (A) {$f_0 \leq 2$};
\node[decision,below=of A, xshift=-2cm, yshift=-0.5cm] (B) {Class-2};
\node[block,right=of B, xshift=2cm,yshift=-1cm] (C) {$f_1\leq 2$};
\node[block, below=of C,xshift=-0.2cm, yshift=-1.6cm] (D) {$f_0 \leq 4$};
\node[decision,right=of D,xshift=2cm, yshift=-2.6cm] (E) {Class-1};
\node[decision, below=of D, xshift=-1cm, yshift=-3cm] (F) {Class-2};
\node[decision, right=of F,xshift=1cm, yshift=-3.5cm] (G) {Class-1};
% \node[decision, below=of D, xshift=2cm,yshift=-2.7cm] (H) {YES};
% \node[decision, left=of H] (I) {NO};
\draw [arrow] (A) -- (B);
\draw [arrow] (A) -- (C);
\draw [arrow] (C) -- (D);
\draw [arrow] (C) -- (E);
\draw [arrow] (D) -- (F);
\draw [arrow] (D) -- (G);

    \end{tikzpicture}
   \caption{Decision tree using the proposed structural impurity (computed using ETC) for Permutation E.}
    \label{figure_dt_sequence_E}   
\end{center}

\end{figure}
\end{comment}

\begin{figure}[!h]
\centerline{ \includegraphics[scale=0.6]{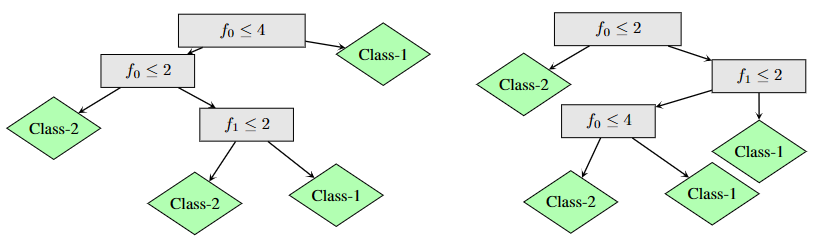}}
\caption{Decision tree using the proposed structural impurity (computed using ETC) for (a) Left: Permutation D, (b) Permutation E.}
\label{figure_dt_sequenceDE} 
\end{figure}

%%%%%%%%%%%%%%%
\end{itemize}
The variability in decision trees obtained from different permutations of data instances (Figures~\ref{fig_toy_example} (b),~\ref{figure_dt_sequenceBC},and~\ref{figure_dt_sequenceDE}, can be attributed to the ETC measure's ability to capture the structural impurity of labels, which was not possible for both Shannon entropy and Gini impurity. Table ~\ref{table_comparison_impurtiy_measures} highlights the sensitivity of ETC to permutation, contrasting with the insensitivity of Shannon entropy and Gini impurity towards data instance permutations. In the given toy example, there are six class-1 data instances and eight class-2 data instances. Since Shannon entropy and Gini impurity are probability-based methods, they remain invariant to label permutation. This sensitivity of ETC to the structural pattern of the label motivates us to develop a bagging algorithm namely \emph{Permutation Decision Forest}.

\begin{table}[!ht]
\centering
\caption{Comparison between Shannon Entropy, Gini Impurity and Structural Impurity (computed using ETC) for the 5 chosen permutations of the data instances of the toy example. Notice the insensitivity of both Shannon entropy and Gini impurity in capturing the changes in the structure of the labels. ETC is able to capture even small changes in structure/ order/ pattern.}
\label{table_comparison_impurtiy_measures}
\scalebox{0.8}{
\begin{tabular}{|c|c|c|c|}
\hline
Label Impurity & \begin{tabular}[c]{@{}c@{}}Shannon \\ Entropy\\ (bits)\end{tabular} & \begin{tabular}[c]{@{}c@{}}Gini\\ Impurity\end{tabular} & \begin{tabular}[c]{@{}c@{}}Structural\\ Impurity (ETC)\end{tabular} \\ \hline
Permutation A & 0.985 & 0.490 & 7 \\ \hline
Permutation B & 0.985 & 0.490 & 8 \\ \hline
Permutation C & 0.985 & 0.490 & 9 \\ \hline
Permutation D & 0.985 & 0.490 & 9 \\ \hline
Permutation E & 0.985 & 0.490 & 8 \\ \hline
\end{tabular}
}
\end{table}

%%%%%
\subsection{Permutation Decision Forest}
Permutation decision forest distinguishes itself from Random Forest by eliminating the need for random subsampling of data and feature selection in order to generate distinct decision trees. Instead, permutation decision forest achieves tree diversity through permutation of the data instances. Since different permutations of the data instances have potentially different structural impurities, we are guaranteed to have a diversity of decision trees by using a large number of permutations. Figure~\ref{fig_pdf_archi} illustrates the operational flow of the proposed permutation decision forest algorithm.

\begin{figure}[!ht]
% \centerline{\includegraphics[width=8cm,height=5.0cm]{method1.png}}
\centerline{ \includegraphics[width=0.7\textwidth]{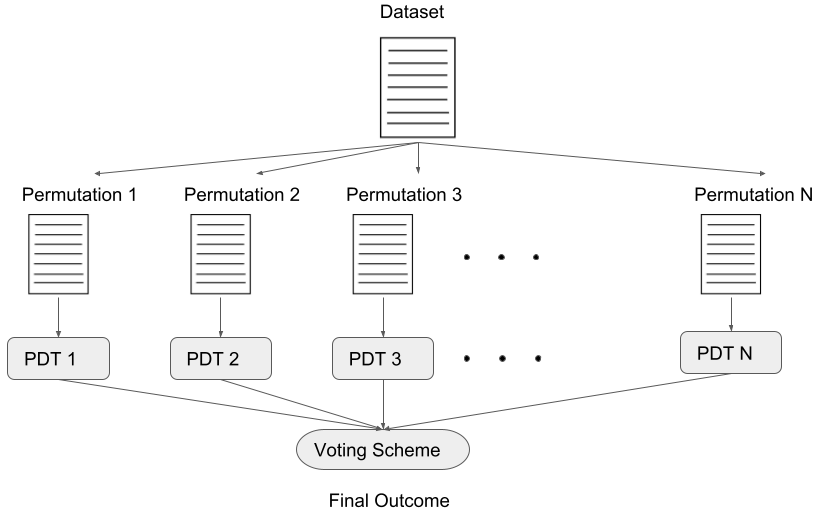}}
\caption{Permutation Decision Forest. The input dataset is subjected to a large number of permutations resulting in different ordering of the data instances. Each such permutation is then used to create a specific permutation decision tree using the structural impurity measure (computed using ETC) as the splitting criteria (ETC gain).  The results from each permutation decision tree are then fed into a majority voting scheme to determine the final predicted label.}
\label{fig_pdf_archi}
\end{figure}

Figure~\ref{fig_pdf_archi} showcases the workflow of the proposed Permutation Decision Forest algorithm, illustrating its functioning. Consisting of individual permutation decision trees, each tree operates on a permuted dataset to construct a classification model, collectively forming a strong classifier. The outcomes of the permutation decision trees are then fed into a majority voting scheme, where the final predicted label is determined by majority votes. Notably, the key distinction between the Permutation Decision Forest and Random Forest lies in their approaches to obtaining distinct decision trees. While Random Forest relies on random subsampling and feature selection, Permutation Decision Forest achieves diversity through permutation of the input data. This is guaranteed by the sensitivity of the structural impurity measure (computed using ETC) to changes in the structure/order/pattern of labels for different permutations of the data instances. Random forest algorithm does not have this benefit since the impurity measure used (Shannon entropy or Gini impurity) is not sensitive to permutations of the data-instances and would yield only one single tree for every permutation. Hence, diversity of trees is impossible to obtain using Random forest algorithm via permutations. This distinction is significant as random feature selection in Random Forest may result in information loss, which is avoided in Permutation Decision Forest. 

%%%%%%%%%%%%%%%
\section{Experiments and Results}
In this section, we discuss about the hyperparamter tuning using five fold crossvalidation and test performance of permutation decision tree, classical decision tree, permutation decision forest and random forest for the following datasets: \emph{Appendicitis}, \emph{Breast Cancer Wisconsin}, \emph{Diabetes Pima Indian}, \emph{Ionosphere}, \emph{Iris}, \emph{Sonar}, and \emph{Wine}. We used sklearn library and python 3 for the implememtation. For all the cases, the dataset was randomly grouped into training and testing. The random seed used for the same is 42. We then do hyperparamter tuning on the training set using timeseries split (a library available in sklearn). Time series split is preferred over $k$ fold crossvalidation in cases where there are temporal order dependencies between data instances. The hyperparameter tuning details are provided in the Appendix section~\ref{section_hyperparameter_tuning} (PDT: Table~\ref{Table_PDT_hyperparameter_tuning}, DT: Table~\ref{Table_DT_hyperparameter_tuning}, PDF: Table~\ref{Table_PDF_hyperparameter_tuning}, and RF: Table~\ref{Table_RF_hyperparameter_tuning}).

\subsection{Performance on Testdata}
We compared the performance of PDT with DT, and PDF with RF on the test data. Figures~\ref{figure_comparison_testdata}(a) and \ref{figure_comparison_testdata}(b) show the respective performance comparisons. In the case of comparison between the performance of PDT with DT as measured by macro f1 score, we observe the following:
\begin{figure}[!h]
    \centering
    \begin{subfigure}[b]{0.49\textwidth}
        \includegraphics[width=\linewidth]{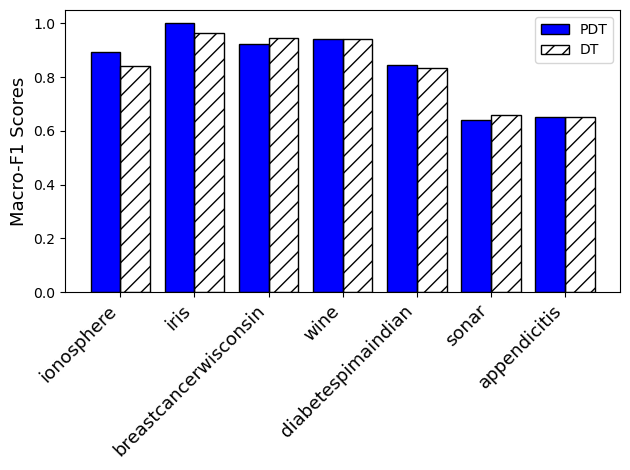} % Replace 'image1.png' with the path to your first image file
       % \caption{Comparative performance evaluation of PDT and DT.}
        %\label{figure_pdt_dt}
    \end{subfigure}
    \hfill
    \begin{subfigure}[b]{0.49\textwidth}
        \includegraphics[width=\linewidth]{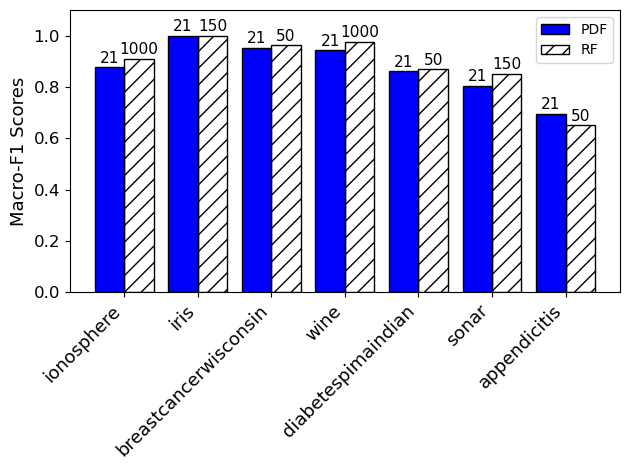} % Replace 'image2.png' with the path to your second image file
        %\caption{Comparative performance evaluation of PDF and RF. The n\_estimators used for PDF and RF are indicated at the top of the bar graph. }
       % \label{figure_pdf_rf}
    \end{subfigure}
    \caption{(a) Left: Comparative performance evaluation of PDT and DT. (b) Right: Comparative performance evaluation of PDF and RF. The n\_estimators used for PDF and RF are indicated at the top of the bar graph. }
    \label{figure_comparison_testdata}
\end{figure}

\begin{itemize}
    \item Figure~\ref{figure_comparison_testdata} (a) shows that PDT slightly outperforms DT on the Ionosphere, Iris, and Diabetes Indian Pima datasets. However, DT slightly outperforms PDT on the Breast Cancer Wisconsin and Sonar datasets. For the Wine and Appendicitis datasets, both PDT and DT perform equally well. The detailed results are provided in Table~\ref{Table_PDT_testdata} and~\ref{Table_DT_Testdata} in the appendix section~\ref{section_hyperparameter_tuning}.
  \item Figure~\ref{figure_comparison_testdata} (b) shows that PDF, created by bagging 21 PDTs, performs comparably to RF, which uses 50 to 1000 trees, across various datasets. Even without random sub-sampling and feature selection, PDF's performance is close to RF. This suggests that further experiments with similar bagging are promising. Detailed results are provided in Table~\ref{Table_PDF_Testdata} and~\ref{Table_RF_Testdata} in the appendix section~\ref{section_hyperparameter_tuning}.

\end{itemize}
%\newpage
\section{Limitations\label{section_limitations}}
The following are the limitations of the proposed research:
\begin{enumerate}
\item Due to computational constraints, we limited our experiments on PDF to 21 n\_estimators. We plan to address this limitation by optimizing the ETC code to ensure a fair comparison with Random Forest.
\item We empirically show that average ETC behaves like Shannon Entropy and hypothesize that individual ETC behaves like Kolmogorov complexity. We aim to rigorously prove this hypothesis in future work.
\item The proposed algorithm is ideal for data with temporal ordering. In future research, we plan to identify such datasets and evaluate the algorithm's effectiveness.
\end{enumerate}
\section{Conclusion}
In this work, we introduced the \emph{Effort-To-Compress} (ETC) complexity measure as a novel impurity measure for decision trees. Unlike traditional measures like Shannon Entropy and Gini impurity, ETC does not assume data instances to be independent and identically distributed. We empirically show that average ETC behaves similarly to Shannon Entropy and hypothesize that individual ETC behaves like Kolmogorov complexity, which we plan to explore further. Additionally, we successfully implemented the proposed \emph{Permutation Decision Tree} (PDT) and \emph{Permutation Decision Forest} (PDF) algorithms on real-world datasets. Our experiments show that PDT outperforms classical decision trees on average, and PDF achieves performance close to Random Forests with significantly fewer estimators (21 vs. 50-1000). Future work will focus on demonstrating a partial order between ETC and Kolmogorov complexity and testing the algorithms on non-iid datasets.

\section*{Acknowledgement}
Harikrishnan N. B. gratefully acknowledges the financial support provided by the New Faculty Seed Grant from BITS Pilani, K. K. Birla Goa Campus. We also extend our gratitude to Dr. Aditya Challa for his insightful discussions on the connection between Shannon entropy and average ETC.

% \section*{References}

% References follow the acknowledgments in the camera-ready paper. Use unnumbered first-level heading for
% the references. Any choice of citation style is acceptable as long as you are
% consistent. It is permissible to reduce the font size to \verb+small+ (9 point)
% when listing the references.
% Note that the Reference section does not count towards the page limit.
% \medskip

% {
% \small

% [1] Alexander, J.A.\ \& Mozer, M.C.\ (1995) Template-based algorithms for
% connectionist rule extraction. In G.\ Tesauro, D.S.\ Touretzky and T.K.\ Leen
% (eds.), {\it Advances in Neural Information Processing Systems 7},
% pp.\ 609--616. Cambridge, MA: MIT Press.

% [2] Bower, J.M.\ \& Beeman, D.\ (1995) {\it The Book of GENESIS: Exploring
%   Realistic Neural Models with the GEneral NEural SImulation System.}  New York:
% TELOS/Springer--Verlag.

% [3] Hasselmo, M.E., Schnell, E.\ \& Barkai, E.\ (1995) Dynamics of learning and
% recall at excitatory recurrent synapses and cholinergic modulation in rat
% hippocampal region CA3. {\it Journal of Neuroscience} {\bf 15}(7):5249-5262.
% }

%%%%%%%%%%%%%%%%%%%%%%%%%%%%%%%%%%%%%%%%%%%%%%%%%%%%%%%%%%%%
\newpage
\appendix

\section{Appendix / supplemental material}
\section{ETC Algorithm\label{sec_ETC_algorithm}}
\begin{algorithm}
    \caption{Finding the most frequently occurring pair in the sequence $S$ (\emph{\textbf{max\_pair\_frequency}}).}
    \textbf{Input:} Sequence $S = s_1,s_2,s_3,\ldots,s_n$, where $s_i \in \{1,2,3,\ldots, K\}$, where $K \in \mathbb{Z}^{+}$.
    
    \textbf{Output:} Return the pair, or set of pairs, in the sequence $S$ that corresponds to the maximum frequency.
    \label{alg_pair_wise_max}
    
    \begin{algorithmic}[1]
        \Require $P \gets \text{zeros}(\max(S),\max(S))$
        \Require $row\_index, col\_index \gets [\,]$
        \For{$i= 1$ \textbf{to} $\text{length}(S)-1$}
            \State $j \gets i+1$
            \State $P(S(i),S(j)) \gets P(S(i),S(j)) +1$
        \EndFor

        \State $max\_val \gets \max(P)$
        \For{$row = 1$ \textbf{to} $\text{size}(P,1)$}
            \For{$col = 1$ \textbf{to} $\text{size}(P,2)$}
                \If{$P(row, col) == max\_val$}
                    \State $row\_index \gets [row\_index, row]$
                    \State $col\_index \gets [col\_index, col]$
                \EndIf
            \EndFor
        \EndFor\\
        \Return $row\_index, col\_index$
    \end{algorithmic}
\end{algorithm}
%%%%%%%%%%%%%%
%\vspace{-5cm}
\begin{algorithm}
    \caption{Finding the first maximum occurring pair in the sequence (\textbf{\emph{pair\_substitution}}).}
    \textbf{Input:} Sequence $S = s_1,s_2,s_3,\ldots,s_n$, where $s_i \in \{1,2,3,\ldots, K\}$, where $K \in \mathbb{Z}^{+}$.
    
    \textbf{Input:} $row\_index, col\_index$
    
    \textbf{Output:} Return the first maximum occurring pair in the sequence.
    \label{alg_first_occuring_max_pair}
    
    \begin{algorithmic}[1]
        \Ensure $substitution\_pair \gets \text{zeros}(1,2)$
        \For{$i= 1$ \textbf{to} $\text{length}(S)-1$}
            \For{$j=1$ \textbf{to} $\text{length}(row\_index)$}
                \If{$(S(i) == row\_index{(j)})$ \& $(S(i+1) == col\_index(j))$ }
                    \State $substitution\_pair \gets [row\_index(j), col\_index(j)]$
                    \State \textbf{break}
                \EndIf
                \If{$(S(i) == row\_index{(j)})$ \& $(S(i+1) == col\_index(j))$ }
                    \State \textbf{break}
                \EndIf
            \EndFor
            \If{$(S(i) == row\_index{(j)})$ \& $(S(i+1) == col\_index(j))$ }
                \State \textbf{break}
            \EndIf
        \EndFor 
        \State \Return $substitution\_pair$
    \end{algorithmic}
\end{algorithm}

\begin{algorithm}[!h]
    \caption{Finding the ETC of the sequence $S$.}
    \textbf{Input:} Sequence $S = s_1,s_2,s_3,\ldots,s_n$, where $s_i \in \{1,2,3,\ldots, K\}$, where $K \in \mathbb{Z}^{+}>0$.
    
    \textbf{Output:} Return the ETC of the sequence $S$.
    \label{alg_ETC_algorithm}
    
    \begin{algorithmic}[1]
        \Ensure $alphabet \gets \{K+1, K+2,\ldots, K+n\}$
        \Ensure $init \gets 1$
        \Ensure $index\_alphabet \gets 1$
        \Ensure $ETC \gets 0$
        \State $[row\_index, col\_index] \gets \text{max\_pair\_frequency}(S)$
        \State $substitution\_pair \gets \text{pair\_substitution}(S, row\_index, col\_index)$
        \State $new\_list \gets S$
        \State $S\_new \gets S$
        \While{$(\text{length}(\text{unique}(S\_new)) \neq 1)$}
            \For{$i = \text{init}:\text{length}(S\_new)-1$}
                \If{$(S\_new(i) == substitution\_pair(1)) \& (S\_new(i+1) == substitution\_pair(2))$}
                    \State $new\_list(i) \gets \text{alphabet}(index\_alphabet)$
                    \State $new\_list(i+1) \gets 0$; $\text{init} \gets i+2$
                \EndIf
            \EndFor
            \State $index\_non\_zero \gets (new\_list \neq 0)$
            \State $ETC \gets ETC + 1$
            \State $S\_new \gets new\_list(index\_non\_zero)$
            \State $new\_list \gets S\_new$
            \State $[row\_index, col\_index] \gets \text{max\_pair\_frequency}(S\_new)$
            \State $substitution\_pair \gets \text{pair\_substitution}(S\_new,row\_index, col\_index)$
            \State $\text{init} \gets 1$
            \State $index\_alphabet \gets index\_alphabet + 1$
        \EndWhile
        \State \textbf{return} \emph{ETC}
    \end{algorithmic}
\end{algorithm}
\newpage
\section{Hyperparameter Tuning\label{section_hyperparameter_tuning}}

To optimize PDT performance, we determine the ideal permutation and depth for each dataset. We explore 21 unique seed values (1 to 21) for shuffling the training set, each generating a distinct dataset permutation. Through timeseries split (five splits, with validation data size = 15) for depths ranging from 2 to 20, we identify the seed value and corresponding depth that produce the highest macro F1-score during hyperparameter tuning. Using the shuffled data and the depth obtained via hyperparameter tuning, we retrain the entire shuffled training dataset.
\subsection{Hyperparameter tuning for classical Decision Tree}
For the classical decision tree, we exclusively adjust the tree depth (using Shannon entropy as the impurity measure) from 2 to 20 via timeseries split (five splits, with a validation data size = 15). We omit shuffling data variations as Shannon entropy remains unaffected by data permutation.
\subsection{Hyperparameter tuning of PDF}
For PDF, we shuffled the data using 21 distinct seed values (1 to 21), resulting in 21 unique permuted datasets. For each permutation, we independently determine the optimal depth (ranging from 2 to 20) via timeseries split (five splits, with a validation data size = 15). During testing, we pass the test data through the 21 PDT models and employ a voting scheme to predict the correct label.
\subsection{Hyperparameter tuning of Random Forest}

For random forest, we employ timeseries split on the training data to optimize n\_estimators and depth. We consider n\_estimators values of 10, 50, 100, 150, and 1000 for tuning. Depth values explored during hyperparameter tuning range from 1 to 21.

% The hyperparameter tuning for all the algorithms are provided in the appendix section.
%%%%%%%%%%%%%%%%%%%%%%%%%%%%%%%%%%%%%%%%%%%%%%%%%%%%%%%%%%%%
\begin{table}[htbp]
\caption{PDT Hyperparameter tuning: The hyperparameters of PDT \emph{Seedvalue} and \emph{Depth} tuned using timeseries split.}
\label{Table_PDT_hyperparameter_tuning}
\centering
\scalebox{0.8}{
\begin{tabular}{|l|c|c|rrrr|}
\hline
                               & \multicolumn{1}{l|}{} & \multicolumn{1}{l|}{} & \multicolumn{4}{c|}{\textbf{Five Fold Cross Validation Mean}}                                                                        \\ \hline
\multicolumn{1}{|c|}{\textbf{Datasets}} & \textbf{Seedvalue}             & \textbf{Depth}                & \multicolumn{1}{c|}{\textbf{Accuracy}} & \multicolumn{1}{c|}{\textbf{F1Score}} & \multicolumn{1}{c|}{\textbf{Precision}} & \multicolumn{1}{c|}{\textbf{Recall}} \\ \hline
Appendicitis                   & 7                     & 4                     & \multicolumn{1}{r|}{0.880}    & \multicolumn{1}{r|}{0.828}   & \multicolumn{1}{r|}{0.840}     & 0.875                       \\ \hline
Breast Cancer Wisconsin        & 2                     & 6                     & \multicolumn{1}{r|}{1.000}    & \multicolumn{1}{r|}{1.000}   & \multicolumn{1}{r|}{1.000}     & 1.000                       \\ \hline
Diabetes Pima Indian           & 20                    & 5                     & \multicolumn{1}{r|}{0.907}    & \multicolumn{1}{r|}{0.899}   & \multicolumn{1}{r|}{0.901}     & 0.906                       \\ \hline
Ionosphere                     & 15                    & 6                     & \multicolumn{1}{r|}{0.920}    & \multicolumn{1}{r|}{0.908}   & \multicolumn{1}{r|}{0.930}     & 0.899                       \\ \hline
Iris                           & 20                    & 2                     & \multicolumn{1}{r|}{0.960}    & \multicolumn{1}{r|}{0.962}   & \multicolumn{1}{r|}{0.967}     & 0.961                       \\ \hline
Sonar                          & 18                    & 4                     & \multicolumn{1}{r|}{0.813}    & \multicolumn{1}{r|}{0.808}   & \multicolumn{1}{r|}{0.816}     & 0.809                       \\ \hline
Wine                           & 21                    & 2                     & \multicolumn{1}{r|}{0.947}    & \multicolumn{1}{r|}{0.946}   & \multicolumn{1}{r|}{0.949}     & 0.953                       \\ \hline
\end{tabular}
}
\end{table}

\begin{table}[!h]
\centering
\caption{DT Hyperparameter Tuning: The hyperparameters of DT - \emph{Depth} tuned using timeseries split.\label{Table_DT_hyperparameter_tuning}}
% \resizebox{\textwidth}{!}{%
\scalebox{0.8}{
\begin{tabular}{|l|c|rrrr|}
\hline
                               & \multicolumn{1}{l|}{} & \multicolumn{4}{c|}{\textbf{DT Cross Validation}}                                                                                         \\ \hline
\multicolumn{1}{|c|}{\textbf{Datasets}} & \textbf{Depth}                 & \multicolumn{1}{c|}{\textbf{Accuracy}} & \multicolumn{1}{c|}{\textbf{F1Score}} & \multicolumn{1}{c|}{\textbf{Precision}} & \multicolumn{1}{c|}{\textbf{Recall}} \\ \hline
Appendicitis                   & 2                     & \multicolumn{1}{r|}{0.800}    & \multicolumn{1}{r|}{0.619}   & \multicolumn{1}{r|}{0.652}     & 0.608                      \\ \hline
Breast Cancer Wisconsin        & 4                     & \multicolumn{1}{r|}{0.947}    & \multicolumn{1}{r|}{0.926}   & \multicolumn{1}{r|}{0.921}     & 0.938                       \\ \hline
Diabetes Pima Indian           & 7                     & \multicolumn{1}{r|}{0.960}    & \multicolumn{1}{r|}{0.946}   & \multicolumn{1}{r|}{0.974}     & 0.933                       \\ \hline
Ionosphere                     & 3                     & \multicolumn{1}{r|}{0.893}    & \multicolumn{1}{r|}{0.879}   & \multicolumn{1}{r|}{0.912}     & 0.885                       \\ \hline
Iris                           & 2                     & \multicolumn{1}{r|}{0.920}    & \multicolumn{1}{r|}{0.918}   & \multicolumn{1}{r|}{0.936}     & 0.924                       \\ \hline
Sonar                          & 5                     & \multicolumn{1}{r|}{0.720}    & \multicolumn{1}{r|}{0.710}   & \multicolumn{1}{r|}{0.718}     & 0.721                       \\ \hline
Wine                           & 4                     & \multicolumn{1}{r|}{0.920}    & \multicolumn{1}{r|}{0.918}   & \multicolumn{1}{r|}{0.916}     & 0.933                       \\ \hline
\end{tabular}
}
\end{table}
%%% TEST %%%%
\begin{table}[!h]
\centering
\caption{PDT Testdata Results.\label{Table_PDT_testdata}}
% \resizebox{\textwidth}{!}{%
\scalebox{0.8}{
\begin{tabular}{|l|c|c|rrrr|}
\hline
                               & \multicolumn{1}{l|}{} & \multicolumn{1}{l|}{} & \multicolumn{4}{c|}{\textbf{Test}}                                       \textbf{}                                                            \\ \hline
\multicolumn{1}{|c|}{\textbf{Datasets}} & \textbf{Seedvalue}             & \textbf{Depth}                 & \multicolumn{1}{c|}{\textbf{Accuracy}} & \multicolumn{1}{c|}{\textbf{F1Score}} & \multicolumn{1}{c|}{\textbf{Precision}} & \multicolumn{1}{c|}{\textbf{Recall}} \\ \hline
Appendicitis                   & 7                     & 4                     & \multicolumn{1}{r|}{0.773}    & \multicolumn{1}{r|}{0.651}   & \multicolumn{1}{r|}{0.728}     & 0.635                       \\ \hline
Breast Cancer Wisconsin        & 2                     & 6                     & \multicolumn{1}{r|}{0.930}    & \multicolumn{1}{r|}{0.925}   & \multicolumn{1}{r|}{0.925}     & 0.925                       \\ \hline
Diabetes Pima Indian           & 20                    & 5                     & \multicolumn{1}{r|}{0.857}    & \multicolumn{1}{r|}{0.844}   & \multicolumn{1}{r|}{0.844}     & 0.844                       \\ \hline
Ionosphere                     & 15                    & 6                     & \multicolumn{1}{r|}{0.901}    & \multicolumn{1}{r|}{0.893}   & \multicolumn{1}{r|}{0.916}     & 0.881                       \\ \hline
Iris                           & 20                    & 2                     & \multicolumn{1}{r|}{1.000}    & \multicolumn{1}{r|}{1.000}   & \multicolumn{1}{r|}{1.000}     & 1.000                       \\ \hline
Sonar                          & 18                    & 4                     & \multicolumn{1}{r|}{0.643}    & \multicolumn{1}{r|}{0.641}   & \multicolumn{1}{r|}{0.656}     & 0.663                       \\ \hline
Wine                           & 21                    & 2                     & \multicolumn{1}{r|}{0.944}    & \multicolumn{1}{r|}{0.942}   & \multicolumn{1}{r|}{0.954}     & 0.935                       \\ \hline
\end{tabular}
}
\end{table}
\begin{table}[!h]
\centering
\caption{DT Testdata Results.\label{Table_DT_Testdata}}
\scalebox{0.8}{
\begin{tabular}{|l|c|rrrr|}
\hline
                               & \multicolumn{1}{l|}{} & \multicolumn{4}{c|}{\textbf{DT Test}}                                                                                         \\ \hline
\multicolumn{1}{|c|}{\textbf{Datasets}} & \textbf{Depth}                 & \multicolumn{1}{c|}{\textbf{Accuracy}} & \multicolumn{1}{c|}{\textbf{F1Score}} & \multicolumn{1}{c|}{\textbf{Precision}} & \multicolumn{1}{c|}{\textbf{Recall}} \\ \hline
Appendicitis                   & 2                     & \multicolumn{1}{r|}{0.773}    & \multicolumn{1}{r|}{0.651}   & \multicolumn{1}{r|}{0.728}     & 0.635                       \\ \hline
Breast Cancer Wisconsin        & 4                     & \multicolumn{1}{r|}{0.947}    & \multicolumn{1}{r|}{0.944}   & \multicolumn{1}{r|}{0.944}     & 0.944                       \\ \hline
Diabetes Pima Indian           & 7                     & \multicolumn{1}{r|}{0.844}    & \multicolumn{1}{r|}{0.833}   & \multicolumn{1}{r|}{0.829}     & 0.838                       \\ \hline
Ionosphere                     & 3                     & \multicolumn{1}{r|}{0.859}    & \multicolumn{1}{r|}{0.843}   & \multicolumn{1}{r|}{0.887}     & 0.828                       \\ \hline
Iris                           & 2                     & \multicolumn{1}{r|}{0.967}    & \multicolumn{1}{r|}{0.966}   & \multicolumn{1}{r|}{0.972}     & 0.963                       \\ \hline
Sonar                          & 5                     & \multicolumn{1}{r|}{0.667}    & \multicolumn{1}{r|}{0.660}   & \multicolumn{1}{r|}{0.661}     & 0.671                       \\ \hline
Wine                           & 4                     & \multicolumn{1}{r|}{0.944}    & \multicolumn{1}{r|}{0.943}   & \multicolumn{1}{r|}{0.958}     & 0.935                       \\ \hline
\end{tabular}
}
\end{table}

%%% TEST %%%%

\begin{table}[!h]
\centering
\caption{PDF Hyperparameter Tuning.\label{Table_PDF_hyperparameter_tuning}}
% \resizebox{\textwidth}{!}{%
\scalebox{0.8}{
\begin{tabular}{|c|l|l|}
\hline
\multicolumn{1}{|l|}{\textbf{Datasets}} & (\textbf{Shuffle seed, Depth})                                                                                                                                                                                                                & \multicolumn{1}{c|}{\textbf{Average Macro F1Score}}                                                                                                                                                                        \\ \hline
Appendicitis                   & \begin{tabular}[c]{@{}l@{}}(1, 2), (2, 3), (3, 2), (4, 2),\\ (5, 3), (6, 2), (7, 4), (8, 2),\\ (9, 2), (10, 3), (11, 2), (12, 2),\\ (13, 2), (14, 2), (15, 2), (16, 3),\\ (17, 3), (18, 2), (19, 2), (20, 5),\\ (21, 2)\end{tabular} & \begin{tabular}[c]{@{}l@{}}0.746, 0.669, 0.686, 0.687,\\ 0.721, 0.669, 0.828, 0.666,\\ 0.728, 0.726, 0.756, 0.731,\\ 0.551, 0.700, 0.648, 0.626,\\ 0.623, 0.623, 0.679, 0.712,\\ 0.720\end{tabular} \\ \hline
Breast Cancer Wisconsin        & \begin{tabular}[c]{@{}l@{}}(1, 4), (2, 6), (3, 8), (4, 4),\\ (5, 2), (6, 4), (7, 2), (8, 2),\\ (9, 3), (10, 4), (11, 3), (12, 2),\\ (13, 3), (14, 6), (15, 3), (16, 2),\\ (17, 3), (18, 2), (19, 3), (20, 2),\\ (21, 3)\end{tabular} & \begin{tabular}[c]{@{}l@{}}0.962, 1.000, 0.971, 0.947,\\ 0.972, 0.939, 0.949, 0.943,\\ 0.898, 0.938, 0.939, 0.932,\\ 0.912, 0.986, 0.883, 0.928,\\ 0.956, 0.938, 0.971, 0.901,\\ 0.919\end{tabular} \\ \hline
Diabetes Pima Indian           & \begin{tabular}[c]{@{}l@{}}(1, 2), (2, 2), (3, 4), (4, 6),\\ (5, 5), (6, 2), (7, 7), (8, 3),\\ (9, 3), (10, 5), (11, 8), (12, 7),\\ (13, 2), (14, 5), (15, 5), (16, 5),\\ (17, 2), (18, 3), (19, 5), (20, 5),\\ (21, 2)\end{tabular} & \begin{tabular}[c]{@{}l@{}}0.861, 0.881, 0.851, 0.814,\\ 0.791, 0.795, 0.808, 0.863,\\ 0.827, 0.809, 0.882, 0.867,\\ 0.865, 0.774, 0.864, 0.768,\\ 0.840, 0.865, 0.766, 0.899,\\ 0.847\end{tabular} \\ \hline
Ionosphere                     & \begin{tabular}[c]{@{}l@{}}(1, 3), (2, 2), (3, 4), (4, 3),\\ (5, 5), (6, 3), (7, 3), (8, 3),\\ (9, 3), (10, 5), (11, 4), (12, 6),\\ (13, 4), (14, 4), (15, 6), (16, 3),\\ (17, 5), (18, 5), (19, 4), (20, 4),\\ (21, 4)\end{tabular} & \begin{tabular}[c]{@{}l@{}}0.877, 0.771, 0.880, 0.834,\\ 0.905, 0.832, 0.886, 0.891,\\ 0.901, 0.887, 0.872, 0.883,\\ 0.854, 0.847, 0.908, 0.812,\\ 0.841, 0.852, 0.884, 0.864,\\ 0.851\end{tabular} \\ \hline
Iris                           & \begin{tabular}[c]{@{}l@{}}(1, 5), (2, 2), (3, 3), (4, 2),\\ (5, 3), (6, 3), (7, 2), (8, 4),\\ (9, 2), (10, 3), (11, 2), (12, 3),\\ (13, 2), (14, 2), (15, 2), (16, 3),\\ (17, 2), (18, 3), (19, 2), (20, 2),\\ (21, 4)\end{tabular} & \begin{tabular}[c]{@{}l@{}}0.946, 0.930, 0.853, 0.957,\\ 0.890, 0.883, 0.908, 0.957,\\ 0.929, 0.931, 0.947, 0.943,\\ 0.919, 0.948, 0.937, 0.914,\\ 0.913, 0.906, 0.942, 0.962,\\ 0.943\end{tabular} \\ \hline
Sonar                          & \begin{tabular}[c]{@{}l@{}}(1, 5), (2, 3), (3, 2), (4, 4),\\ (5, 4), (6, 4), (7, 4), (8, 4),\\ (9, 3), (10, 3), (11, 4), (12, 3),\\ (13, 4), (14, 3), (15, 5), (16, 5),\\ (17, 2), (18, 4), (19, 3), (20, 5),\\ (21, 5)\end{tabular} & \begin{tabular}[c]{@{}l@{}}0.742, 0.751, 0.718, 0.762,\\ 0.671, 0.696, 0.741, 0.689,\\ 0.677, 0.702, 0.694, 0.602,\\ 0.731, 0.789, 0.700, 0.743,\\ 0.651, 0.808, 0.686, 0.707,\\ 0.657\end{tabular} \\ \hline
Wine                           & \begin{tabular}[c]{@{}l@{}}(1, 2), (2, 2), (3, 3), (4, 2),\\ (5, 4), (6, 2), (7, 3), (8, 3),\\ (9, 2), (10, 2), (11, 3), (12, 2),\\ (13, 2), (14, 2), (15, 5), (16, 2),\\ (17, 2), (18, 2), (19, 2), (20, 2),\\ (21, 2)\end{tabular} & \begin{tabular}[c]{@{}l@{}}0.901, 0.827, 0.888, 0.930,\\ 0.858, 0.902, 0.940, 0.932,\\ 0.836, 0.838, 0.875, 0.944,\\ 0.853, 0.918, 0.911, 0.904,\\ 0.895, 0.852, 0.913, 0.893,\\ 0.946\end{tabular} \\ \hline
\end{tabular}}
\end{table}

\begin{table}[!h]
\centering
\caption{RF Hyperparameter Tuning.\label{Table_RF_hyperparameter_tuning}}
% \resizebox{\textwidth}{!}{%
\scalebox{0.8}{
\begin{tabular}{|l|c|c|cccc|}
\hline
\multicolumn{1}{|c|}{}  & \multicolumn{1}{l|}{}           & \multicolumn{1}{l|}{}      & \multicolumn{4}{c|}{\textbf{RF Cross Validation}}                                                                           \\ \hline
\textbf{Datasets }               & \multicolumn{1}{l|}{\textbf{Nestimator}} & \multicolumn{1}{l|}{\textbf{Depth}} & \multicolumn{1}{c|}{\textbf{Accuracy}} & \multicolumn{1}{c|}{\textbf{F1Score}} & \multicolumn{1}{c|}{\textbf{Precision}} & \textbf{Recall} \\ \hline
Appendicitis            & 50                              & 3                          & \multicolumn{1}{c|}{0.867}    & \multicolumn{1}{c|}{0.693}   & \multicolumn{1}{c|}{0.792}     & 0.727  \\ \hline
Breast Cancer Wisconsin & 50                              & 4                          & \multicolumn{1}{c|}{0.973}    & \multicolumn{1}{c|}{0.958}   & \multicolumn{1}{c|}{0.958}     & 0.958  \\ \hline
Diabetes Pima Indian    & 50                              & 11                         & \multicolumn{1}{c|}{0.947}    & \multicolumn{1}{c|}{0.944}   & \multicolumn{1}{c|}{0.951}     & 0.944  \\ \hline
Ionosphere              & 1000                            & 4                          & \multicolumn{1}{c|}{0.933}    & \multicolumn{1}{c|}{0.922}   & \multicolumn{1}{c|}{0.948}     & 0.919  \\ \hline
Iris                    & 150                             & 3                          & \multicolumn{1}{c|}{0.947}    & \multicolumn{1}{c|}{0.945}   & \multicolumn{1}{c|}{0.954}     & 0.952  \\ \hline
Sonar                   & 150                             & 8                          & \multicolumn{1}{c|}{0.813}    & \multicolumn{1}{c|}{0.812}   & \multicolumn{1}{c|}{0.816}     & 0.825  \\ \hline
Wine                    & 1000                            & 2                          & \multicolumn{1}{c|}{0.987}    & \multicolumn{1}{c|}{0.985}   & \multicolumn{1}{c|}{0.983}     & 0.990  \\ \hline
\end{tabular}}
\end{table}
%\vspace{-33pt}

\begin{table}[!h]
\centering
\caption{PDF Testdata Results.\label{Table_PDF_Testdata}}
% \resizebox{\textwidth}{!}{%
\scalebox{0.8}{
\begin{tabular}{|l|cccc|}
\hline
\multicolumn{1}{|c|}{}  & \multicolumn{4}{|c|}{\textbf{PDF Test}}                                                                          \\ \hline
\textbf{Datasets}                & \multicolumn{1}{c|}{\textbf{Accuracy}} & \multicolumn{1}{c|}{\textbf{F1Score}} & \multicolumn{1}{c|}{\textbf{Precision}} & \textbf{Recall} \\ \hline
Appendicitis            & \multicolumn{1}{c|}{0.818}    & \multicolumn{1}{c|}{0.694}   & \multicolumn{1}{c|}{0.900}     & 0.667  \\ \hline
Breast Cancer Wisconsin & \multicolumn{1}{c|}{0.956}    & \multicolumn{1}{c|}{0.953}   & \multicolumn{1}{c|}{0.955}     & 0.951  \\ \hline
Diabetes Pima Indian    & \multicolumn{1}{c|}{0.870}    & \multicolumn{1}{c|}{0.862}   & \multicolumn{1}{c|}{0.856}     & 0.871  \\ \hline
Ionosphere              & \multicolumn{1}{c|}{0.887}    & \multicolumn{1}{c|}{0.877}   & \multicolumn{1}{c|}{0.906}     & 0.863  \\ \hline
Iris                    & \multicolumn{1}{c|}{1.000}    & \multicolumn{1}{c|}{1.000}   & \multicolumn{1}{c|}{1.000}     & 1.000  \\ \hline
Sonar                   & \multicolumn{1}{c|}{0.810}    & \multicolumn{1}{c|}{0.806}   & \multicolumn{1}{c|}{0.805}     & 0.822  \\ \hline
Wine                    & \multicolumn{1}{c|}{0.944}    & \multicolumn{1}{c|}{0.942}   & \multicolumn{1}{c|}{0.954}     & 0.935  \\ \hline
\end{tabular}}
\end{table}

\begin{table}[!h]
\centering
\caption{RF Testdata Results.\label{Table_RF_Testdata}}
% \resizebox{\textwidth}{!}{%
\scalebox{0.8}{
\begin{tabular}{|l|c|c|cccc|}
\hline
\multicolumn{1}{|c|}{}  & \multicolumn{1}{l|}{}           & \multicolumn{1}{l|}{}      & \multicolumn{4}{c|}{\textbf{RF Test}}                                                                           \\ \hline
\textbf{Datasets }               & \multicolumn{1}{l|}{\textbf{Nestimator}} & \multicolumn{1}{l|}{\textbf{Depth}} & \multicolumn{1}{c|}{\textbf{Accuracy}} & \multicolumn{1}{c|}{\textbf{F1Score}} & \multicolumn{1}{c|}{\textbf{Precision}} & \textbf{Recall} \\ \hline
Appendicitis            & 50                              & 3                          & \multicolumn{1}{c|}{0.773}    & \multicolumn{1}{c|}{0.651}   & \multicolumn{1}{c|}{0.728}     & 0.635  \\ \hline
Breast Cancer Wisconsin & 50                              & 4                          & \multicolumn{1}{c|}{0.965}    & \multicolumn{1}{c|}{0.962}   & \multicolumn{1}{c|}{0.967}     & 0.958  \\ \hline
Diabetes Pima Indian    & 50                              & 11                         & \multicolumn{1}{c|}{0.877}    & \multicolumn{1}{c|}{0.868}   & \multicolumn{1}{c|}{0.863}     & 0.876  \\ \hline
Ionosphere              & 1000                            & 4                          & \multicolumn{1}{c|}{0.915}    & \multicolumn{1}{c|}{0.909}   & \multicolumn{1}{c|}{0.926}     & 0.899  \\ \hline
Iris                    & 150                             & 3                          & \multicolumn{1}{c|}{1.000}    & \multicolumn{1}{c|}{1.000}   & \multicolumn{1}{c|}{1.000}     & 1.000  \\ \hline
Sonar                   & 150                             & 8                          & \multicolumn{1}{c|}{0.857}    & \multicolumn{1}{c|}{0.852}   & \multicolumn{1}{c|}{0.847}     & 0.861  \\ \hline
Wine                    & 1000                            & 2                          & \multicolumn{1}{c|}{0.972}    & \multicolumn{1}{c|}{0.976}   & \multicolumn{1}{c|}{0.978}     & 0.976  \\ \hline
\end{tabular}}

\end{table}

\newpage

\end{document}